\documentclass[10pt,twocolumn,letterpaper]{article}

\usepackage{cvpr}              

\usepackage[table]{xcolor}
\usepackage{multirow}
\usepackage{algorithm}
\usepackage{algorithmic}
\usepackage{amsmath}
\usepackage[T1]{fontenc}
\usepackage[table]{xcolor}
\usepackage{soul}
\usepackage[accsupp]{axessibility}

\definecolor{cvprblue}{rgb}{0.21,0.49,0.74}
\usepackage[pagebackref,breaklinks,colorlinks,allcolors=cvprblue]{hyperref}

\newcommand{\hlo}[1]{{\sethlcolor{orange!15}\hl{#1}}}      
\newcommand{\hlb}[1]{{\sethlcolor{cyan!30}\hl{#1}}}   

\def\paperID{40176} 
\def\confName{CVPR}
\def\confYear{2026}

\title{CD-Buffer: Complementary Dual-Buffer Framework for Test-Time Adaptation in Adverse Weather Object Detection}

\author{
Youngjun Song$^{1}$ \quad Hyeongyu Kim$^{1}$ \quad Dosik Hwang$^{1,2 \dagger}$  \\
$^{1}$Yonsei University \quad $^{2}$ Korea Institute of Science and Technology\\
{\tt\small \{yjuuun, lion4309, dosik.hwang\}@yonsei.ac.kr}\\
\small\url{https://wkfsksdl99.github.io/cd_buffer/}
}

\begin{document}
\maketitle{}
\begin{abstract}
Test-Time Adaptation (TTA) enables real-time adaptation to domain shifts without off-line retraining. Recent TTA methods have predominantly explored additive approaches that introduce lightweight modules for feature refinement. Recently, a subtractive approach that removes domain-sensitive channels has emerged as an alternative direction. We observe that these paradigms exhibit complementary effectiveness patterns: subtractive methods excel under severe shifts by removing corrupted features, while additive methods are effective under moderate shifts requiring refinement. However, each paradigm operates effectively only within limited shift severity ranges, failing to generalize across diverse corruption levels. This leads to the following question: can we adaptively balance both strategies based on measured feature-level domain shift?
We propose \textbf{CD-Buffer, a novel complementary dual-buffer framework} where subtractive and additive mechanisms operate in opposite yet coordinated directions driven by a unified discrepancy metric. Our key innovation lies in the discrepancy-driven coupling: Our framework couples removal and refinement through a unified discrepancy metric, automatically balancing both strategies based on feature-level shift severity. This establishes automatic channel-wise balancing that adapts differentiated treatment to heterogeneous shift magnitudes without manual tuning. Extensive experiments on KITTI, Cityscapes, and ACDC datasets demonstrate state-of-the-art performance, consistently achieving superior results across diverse weather conditions and severity levels.
\end{abstract}

\renewcommand*{\thefootnote}{}
\footnotetext[1]{$\dagger$ Corresponding author.} 
\section{Introduction}
\label{sec:introduction}
Object detection (OD) is a fundamental task in computer vision that classifies objects and localizes them with bounding boxes \cite{ren2015faster}. Recent advances in deep learning have significantly improved OD performance, enabling applications in robotics \cite{zhang2024uni}, autonomous driving \cite{dong2023benchmarking}, security, medical imaging \cite{medicalOD}, and logistics \cite{mittal2024comprehensive}. In particular, AI-based object detection remains central to autonomous driving systems and continues to attract active research attention \cite{caesar2020nuscenes}.

However, real-world driving scenarios introduce a critical challenge: domain shift caused by weather variations can drastically degrade detector performance when models are trained only on clear-weather data \cite{bijelic2020seeing}. Domain shift arises when the test data distribution diverges from the training distribution \cite{fang2024source}. Adverse conditions such as rain, snow, and fog cause scattering, occlusions, and reduced visibility, which corrupt image quality and widen the domain gap \cite{bijelic2020seeing}. Consequently, detectors trained primarily on clear-weather datasets suffer from substantial performance drops under these conditions \cite{ACDC, sakaridis2018model, zhao2024revisiting}.

To address such degradation, Domain Adaptation (DA) methods employ transfer or unsupervised learning strategies to align feature distributions \cite{ganin2016domain, wilson2020survey}. However, these methods typically require pre-collected target data \cite{chen2018domain}, extensive retraining \cite{yoo2022unsupervised}, or even target annotations \cite{chen2021semi}, which are impractical for dynamic real-world applications like autonomous driving. Test-Time Adaptation (TTA) provides a more practical alternative by updating source-trained models on-line using only test streams, eliminating the need for labeled target data or off-line retraining \cite{choi2022improving}.

Most existing TTA methods fine-tune batch normalization (BN) layers to adjust feature statistics through affine parameter updates \cite{BNtuning_lee2024entropy,BNtuning_li2018adaptive,BNtuning_niu2022efficient,BNtuning_wang2020tent,BNtuning_yuan2023robust,BNtuning_zhao2023delta}. Although computationally efficient, these methods depend heavily on batch size, leading to instability under small-batch conditions. To overcome this, recent works have explored structural modifications. Additive approaches introduce lightweight adapter modules to learn target-specific refinements while preserving source knowledge  \cite{kim2025buffer, shin2024tta, song2023ecotta, yoo2024WHW}, whereas subtractive approaches prune domain-sensitive channels to focus adaptation on robust features without adding parameters \cite{wang2025pruning}.

\begin{figure}[t]
\centering
\includegraphics[width=\columnwidth]{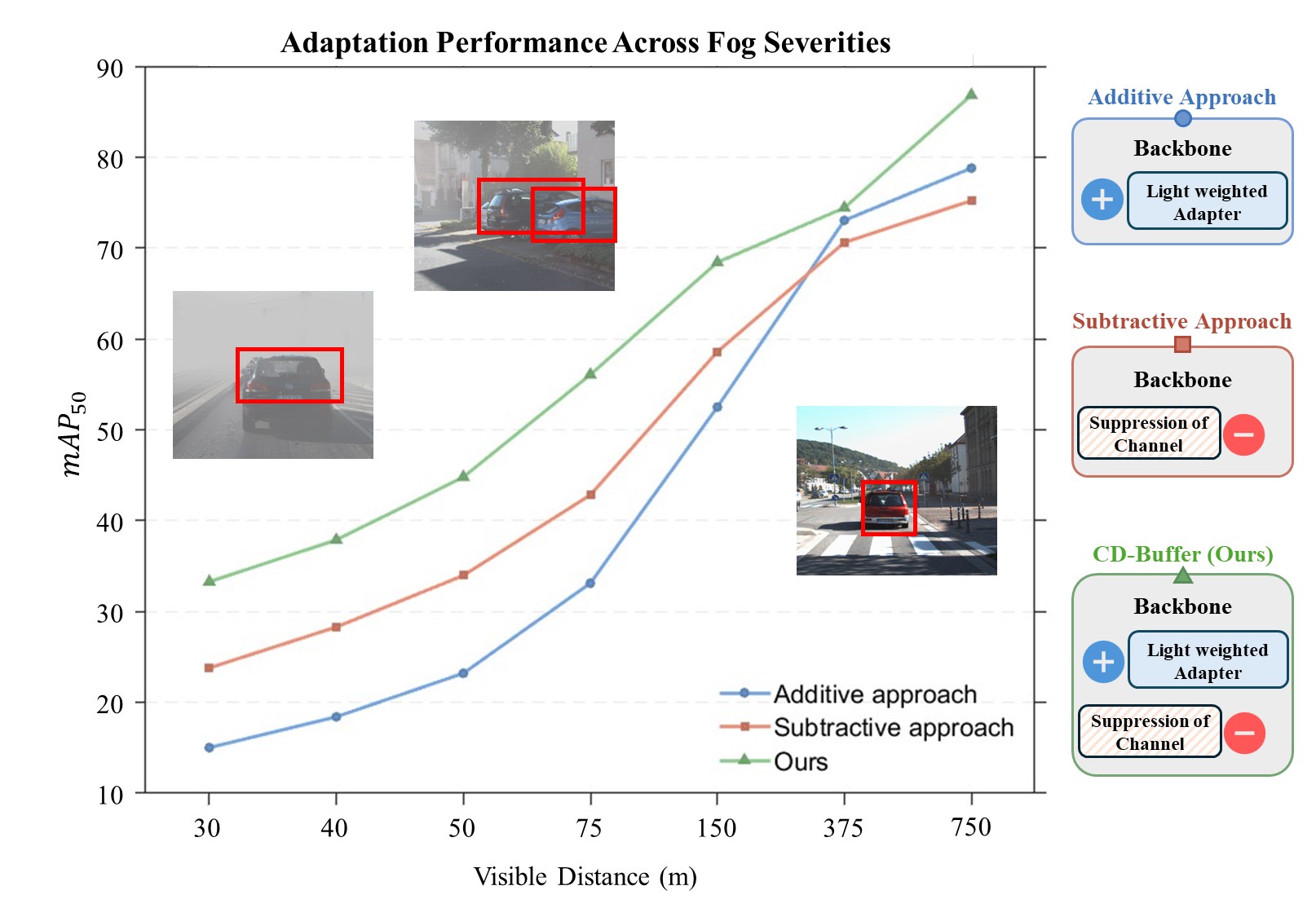}
\caption{\textbf{Adaptation Performance Across Fog Severities.} Additive method \cite{kim2025buffer} excels under moderate shifts but struggle under severe corruption, while subtractive method \cite{wang2025pruning} shows the opposite pattern. Our CD-Buffer achieves consistent performance across all severities by adaptively balancing both strategies.}
\label{fig1.motivation}
\end{figure}

These two paradigms exhibit distinct behaviors depending on the degree of domain shift. Our preliminary analysis reveals distinct effectiveness patterns: under severe domain shifts where features are heavily corrupted, subtractive methods prove effective by removing detrimental channels, while additive methods less effective as refinement modules struggle to correct severely degraded features. Conversely, under moderate shifts where features require subtle adjustment, additive methods excel through fine-grained refinement, whereas subtractive methods risk removing channels that could benefit from correction rather than removal. This observation highlights a core limitation: no single approach consistently performs well across varying shift severities in real-world conditions.

Motivated by this, we introduce CD-Buffer, a novel complementary dual-buffer framework where adaptation strategy is determined individually for each channel based on measured domain shift severity. Our approach measures channel-wise discrepancy from source statistics and adapts accordingly through two coordinated mechanisms. The subtractive buffer employs learnable mask scores to suppress severely shifted channels, while the additive buffer provides compensation through lightweight adapters, with the compensation strength inversely proportional to the degree of suppression. This mechanism enables channel-wise differentiation: severely shifted features are suppressed and compensated, moderately shifted features are refined, and stable features remain largely unaffected, all driven by a unified discrepancy metric. Our contributions are summarized as follows:
\begin{itemize}

\item We reveal that additive and subtractive TTA paradigms are complementary: subtractive methods are more effective under severe shifts, while additive methods perform better under moderate shifts.

\item We propose a discrepancy-driven dual-buffer framework that automatically balances channel suppression and compensation through a shared discrepancy metric.

\item Extensive experiments on adverse weather detection benchmarks demonstrate that our method achieves state-of-the-art performance across various severities.
\end{itemize}
\section{Related Work}
\label{sec:related work}

\subsection{Object Detection}
OD aims to simultaneously predict object categories and localize them with bounding boxes. Modern detectors are broadly categorized into two-stage and one-stage architectures. Two-stage detectors like Faster R-CNN \cite{ren2015faster} employ a cascaded structure where a Region Proposal Network (RPN) first generates candidate regions followed by a classifier that predicts categories and refines coordinates. While achieving high accuracy through this sequential processing with anchor-based proposals, computational bottlenecks remain due to the multi-stage pipeline. One-stage detectors \cite{yolo, redmon2018yolov3, li2022yolov6} address this by performing detection and classification simultaneously through a unified pipeline, achieving significant speedup at the cost of accuracy trade-offs. Our work builds upon Faster R-CNN as the base detector architecture.

\subsection{Test-Time Adaptation}
Test-time adaptation addresses domain shift between source and target domains without access to target labels or off-line retraining. Research focuses on two key aspects: where to adapt in the model and how to optimize without supervision.

Full-parameter adaptation methods like AdaContrast \cite{chen2022contrastive} and CoTTA \cite{wang2022continual} update entire model weights through self-supervised learning or teacher-student consistency losses with stochastic weight restoration to mitigate catastrophic forgetting. While achieving strong adaptation by leveraging full model capacity, these approaches incur substantial computational and memory costs, limiting practicality under real-time constraints.

Normalization-based adaptation offers a more efficient alternative by adjusting only batch normalization layers, based on the hypothesis that domain-related knowledge resides in BN statistics \cite{BNtuning_wang2020tent}. AdaBN \cite{BNtuning_li2018adaptive} recalculates BN statistics from target batches without parameter optimization, while TENT \cite{BNtuning_wang2020tent} optimizes BN affine parameters through entropy minimization, achieving adaptation with minimal memory overhead. However, these Normalization-based methods rely on batch statistics estimation, leading to performance degradation with small batch sizes.

Recent work also explores architectural modifications through additive and subtractive approaches. Additive approaches introduce lightweight modules while freezing pre-trained weights \cite{shin2024tta, song2023ecotta, kim2025buffer, yoo2024WHW}. L-TTA \cite{shin2024tta} inserts lightweight CNN layers at the network input, EcoTTA \cite{song2023ecotta} adds parallel meta-networks to each block, and Buffer Layer \cite{kim2025buffer} inserts modules in residual paths. These approaches preserve source knowledge and avoid batch-size dependency but introduce additional parameters.

Conversely, subtractive approaches remove detrimental components \cite{wang2025pruning, ma2025surgeon}. Wang et al. \cite{wang2025pruning} observed that domain-sensitive channels can negatively impact performance, proposing channel pruning to suppress these channels while concentrating adaptation on robust features. While achieving efficiency without added parameters, this risks information loss and struggles under moderate shifts where features could benefit from refinement rather than removal.

These complementary paradigms exhibit a fundamental characteristic: their relative effectiveness varies with domain shift magnitude. Additive methods excel when features require subtle adjustments, while subtractive methods prove superior when features are severely corrupted.

\subsection{Test-Time Adaptation for Object Detection}
OD poses unique TTA challenges by requiring simultaneous localization and classification, necessitating specialized adaptation strategies.

ActMAD \cite{mirza2023actmad} estimates feature distributions across multiple network layers, aligning each with source statistics through L1 regularization while updating all parameters, enabling location-aware alignment. WHW \cite{yoo2024WHW} introduces lightweight adapter modules in parallel to backbone blocks, updating only adapters while considering both image-level and object-level feature distributions. PruningTTA \cite{wang2025pruning} propose a subtractive approach, pruning domain-sensitive channels measured at both image and object levels with stochastic reactivation to prevent inadvertent removal of useful channels. While achieving efficiency and strong performance on adverse weather detection, this method faces risks of information loss.

These methods exhibit complementary trade-offs. Additive approaches preserve knowledge and provide fine-grained refinement but show limited effectiveness under severe shifts, as they may struggle to recover severely corrupted features. Subtractive approaches are effective under severe shifts by removing corrupted channels but risk information loss and are less effective under moderate shifts, where features would benefit more from refinement than removal.
Critically, existing methods rely on a single paradigm, achieving strong performance only within specific shift severity ranges. This limitation is problematic because domain shifts exhibit varying magnitudes even within a single scene: some feature channels may be severely corrupted while others remain largely intact. Yet no existing method addresses this heterogeneity explicitly. They apply uniform treatment across all channels regardless of individual shift severity. This observation motivates our unified framework that dynamically balances both strategies through channel-wise adaptive modulation, enabling differentiated treatment based on measured feature-level domain shift.

\section{Methods}
\label{sec:methods}
    \begin{figure*}
        \centering
        \includegraphics[width=\textwidth]{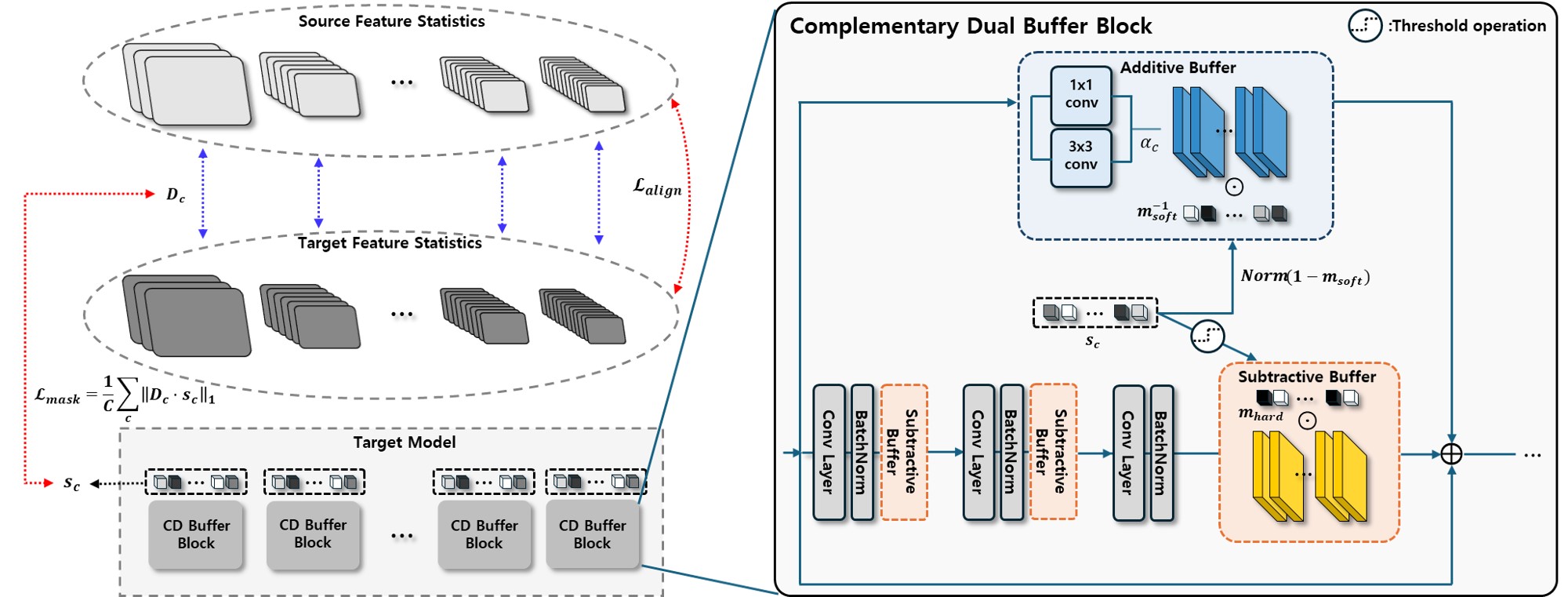}
        \caption{\textbf{Overview of the proposed CD-Buffer framework.} We compute feature-level domain discrepancy $D_c$ by comparing target and pre-computed source statistics. This metric drives two complementary buffers: the \textit{Subtractive Buffer} suppresses severely shifted channels via discrepancy-weighted regularization on learnable mask scores, while the \textit{Additive Buffer} provides adaptive compensation through lightweight convolutions with inverse soft masking, where aggressively suppressed channels receive stronger compensation. Both buffers are unified by feature alignment loss, enabling differentiated adaptation where severely shifted features are both suppressed and compensated, while stable features undergo minimal intervention.}
        \label{fig2.framework}
    \end{figure*}
\subsection{Overview}
We propose the complementary dual-buffer (CD-Buffer) framework, where subtractive and additive mechanisms are driven by the same discrepancy metric in opposite directions. The framework leverages a unified domain discrepancy metric to adaptively modulate features at the channel level. Channels with severe domain shifts are suppressed by the subtractive buffer while simultaneously receiving strong compensation from the additive buffer through inverse soft masking. Channels with moderate shifts undergo balanced refinement, and stable channels experience minimal intervention. This discrepancy-driven modulation operates in a channel-wise adaptive manner, enabling robust adaptation across heterogeneous shift severities.

Figure \ref{fig2.framework} illustrates our framework architecture. We compute domain discrepancy by comparing target feature statistics with pre-computed source statistics. This metric drives two complementary buffers: the subtractive buffer suppresses severely shifted channels via learnable mask scores, while the additive buffer provides adaptive compensation with strength inversely proportional to suppression extent. Both mechanisms are unified through feature alignment optimization, creating automatic balancing between removal and refinement based on measured feature-level domain shift.

\subsection{Feature-Level Domain Discrepancy Score}
To quantify the degree of feature-level domain shift, we measure distance between source and target features. 
Following prior work \cite{wang2025pruning}, we compute this feature at both image-level and instance-level to capture comprehensive domain shifts relevant for object detection tasks.

During source training, we pre-compute source feature statistics. at each input of BN layer $l$,  
mean of source image features $\bar{X}_s^c\in \mathbb{R}^{C \times H \times W}$ and mean of instance feature $\bar{x}_s^c \in \mathbb{R}^{C \times h \times w}$ computed by RoI alignment operation. where $H \times W$, $h \times w$ are feature dimensions.
At test time, we extract corresponding target features $X_t^c\in \mathbb{R}^{C \times H \times W}$ and $x_t^c\in \mathbb{R}^{C \times h \times w}$ from per batch.
The feature-level domain discrepancy for channel $c$ combines both levels:
\begin{equation}
D^I = \frac{\sum_{n=1}^{N} \lVert{X}_t^c - \bar{X}_s^c\rVert_1}{N H W}
\label{eq1}
\end{equation}
\begin{equation}
D^O = \frac{\sum_{m=1}^{M} \lVert{x}_t^c - \bar{x}_s^c\rVert_1}{M h w}
\label{eq2}
\end{equation}

Then $D^I$ and $D^O$ are normalized and subsequently added as shown in the equation below

\begin{equation}
D = D^I + D^O
\label{eq3}
\end{equation}

where $N$ is batch size, $M$ is number of instances.
Higher $D$ indicates severe channel-specific shift requiring intervention, while lower $D$ suggests a moderate shift. 

\subsection{Subtractive Buffer Module}
The subtractive buffer eliminates features that exhibit severe domain shift, which would otherwise significantly degrade model performance. Based on the measured feature-level domain discrepancy $D$, we selectively remove corrupted features while preserving stable features.

We introduce learnable mask scores $s \in \mathbb{R}^C$ parameterized with learnable parameter $\theta$ and initialized from BN weight magnitudes $s_c = |\gamma_c|$. The forward pass applies channel-suppression to after batch normalized features:
\begin{equation}
F^{s}_{\text{out}} = \text{diag}(m_{\text{hard}}) \cdot \text{BN}(F; \gamma, \beta)
\label{eq4}
\end{equation}
where $m_{hard} \in \{0,1\}^C$ is a binary mask determined by the mask scores $s$, $F$ is input feature map, $\gamma, \beta$ are BN affine parameters.

\noindent\textbf{Discrepancy-Weighted Regularization.} The key mechanism linking domain discrepancy to channel suppression is a weighted regularization applied to mask scores:
\begin{equation}
\mathcal{L}_{\text{mask}} = \frac{1}{C} \sum_c \lVert D_{c} \cdot s_{c} \rVert_1
\label{eq5}
\end{equation}
This creates direct coupling: channels with high discrepancy $D_c$ receive stronger gradients to decrease their mask scores $s_c$, promoting sparsity and eventual suppression. Channels with low discrepancy experience weak regularization, allowing their scores to remain high and preserving the channels.

\noindent\textbf{Feature Suppression Mask Generation.}The binary mask is determined by thresholding the learned scores:
\begin{equation}
m_{hard} =
\begin{cases}
1, & \text{if } |s_c| \geq \tau \\
0, & \text{otherwise}
\end{cases}
\label{eq6}
\end{equation}

where the threshold $\tau$ is computed dynamically from the network-wide score distribution:
\begin{equation}
\tau = \text{Percentile}\big(\{|s_c^{(l)}| : \forall l, c\}, \rho_{\text{target}}\big)
\label{eq7}
\end{equation}
This ensures precise control over the channel suppression ratio $\rho_{\text{target}}$.
To enable gradient flow through the discrete binary mask, we employ a reparameterization trick. We compute a hard binary mask $m_{\text{hard}}$ through equation \ref{eq6}, and a differentiable soft mask $m_{\text{soft}} = \sigma((|s| - \tau)/\lambda_S)$, then combine via:
\begin{equation}
m = m_{\text{hard}} + (m_{\text{soft}} - \text{sg}(m_{\text{soft}}))
\label{eq8}
\end{equation}
where $\sigma(\cdot)$ represents sigmoid function with temperature parameter $\lambda_S$, $\text{sg}(\cdot)$ denotes stop-gradient operation.

Additionally, we apply stochastic reactivation to suppressed channels, preventing the permanent removal of potentially useful ones.

\subsection{Additive Buffer Module}
To realign shifted feature distributions and compensate for information lost through channel suppression, we introduce an additive buffer composed of lightweight learnable adapters. Each adapter consists of parallel $1\times1$ and $3\times3$ convolutions that refine intermediate features and inject them via residual connections. Critically, the adaptation strength of these modules is inversely modulated by the same channel-wise discrepancy metric $D$ used in the subtractive buffer, serving two complementary purposes: (1) providing strong compensation for aggressively suppressed channels to recover lost information, and (2) adaptively scaling refinement intensity according to the degree of feature shift. Severely shifted channels receive substantial adjustment, while stable channels undergo minimal modification. This dual functionality enables automatic balancing between removal and compensation without manual tuning.
\begin{equation}
F_{\text{add}} = \frac{\text{Conv}_{1\times1}(F) + \text{Conv}_{3\times3}(F)}{2} \odot \boldsymbol{\alpha}
\label{eq9}
\end{equation}
The out features are added to the residual path with learnable channel-wise scaling factor  $\boldsymbol{\alpha} \in \mathbb{R}^C$ initialized near zero ($10^{-2}$) to preserve source knowledge initially.
The key innovation is inverse soft masking that links both buffers through the same mask scores. 
We compute a soft probabilistic mask $\hat{m}_{\text{soft}} \in \mathbb{R}^c$ from the subtractive buffer's mask scores $s$:
\begin{equation}
\hat{m}_{\text{soft}} = \sigma\!\left(\frac{|s| - \tau}{\lambda_A}\right)
\label{eq10}
\end{equation}

\begin{equation}
    \hat{m}^{-1}_{\text{soft}} = k \cdot Norm(\textbf{1} - \hat{m}_{\text{soft}})
\label{eq11}
\end{equation}

\begin{equation}
F^{\text{a}}_{\text{out}} =  \text{diag}(\boldsymbol{\hat{m}^{-1}_{\text{soft}}}) \cdot F_{\text{add}}
\label{eq12}
\end{equation}
with smoother temperature $\lambda_A > \lambda_S$ and $k$ serves as a scaling parameter that controls the range of additive modulation. This mechanism enables implicit coordination between subtraction and compensation, allowing both buffers to adaptively respond to the degree of feature shift based on the discrepancy metric.

\subsection{Optimization Strategy}
The optimization objective combines feature alignment with conditional masking regularization.
We minimize statistical divergence between target and source distributions:
\begin{equation}
\mathcal{L}_{\text{align}} = 
\sum_l \left(
\lVert\mu_t^l - \bar{\mu}_s^l \rVert _1 + 
\lVert{\sigma}_t^l - \bar{{\sigma}}_s^l \rVert _1
\right)
\label{eq13}
\end{equation}
Here, $\mu$ and $\sigma$ denote the mean and variance of the feature distributions, respectively, and $t$ and $s$ indicate the target and source domains. The total loss is
\begin{equation}
\mathcal{L} = \mathcal{L}_{\text{align}} + 
\lambda_{\text{reg}} \cdot 
\mathcal{L}_{\text{mask}}
\label{eq14}
\end{equation}
where $\lambda_{reg}$ is balance parameter.

Beyond channel-level modulation, we employ layer-level gradient scaling that concentrates computational effort on severely shifted layers. For this layer-level adaptation, we aggregate channel discrepancies:
\begin{equation}
D^l = \frac{1}{C} \sum_c D^{l}_c
\label{eq15}
\end{equation}
After backpropagation but before the optimizer step, we amplify additive buffer gradients proportionally to layer discrepancy $D^l$. We jointly optimize three sets of parameters: the additive buffer parameters, the BN affine parameters $(\gamma, \beta)$, and the mask scores $\mathbf{s}$.

\section{Experiments}
\label{sec:Experiments}

\subsection{Datasets}
\textbf{Source Domain.} We train source models on two large-scale clear-weather driving datasets: KITTI \cite{kittidataset} and Cityscapes \cite{cordts2016cityscapes}. KITTI comprises 7,481 training images and 7,518 test images with 8 object categories. Cityscapes provides 2,975 training images, 500 validation images, and 1,525 test images (without labels) across 8 categories. Both datasets capture clean daytime driving conditions.

\noindent{\textbf{Target Domain.}} For test-time adaptation, we construct adverse weather test sets using physics-based rendering. Following Halder et al. \cite{halder2019physics}, we synthesize fog and rain effects on KITTI test images, creating KITTI-Fog with three visibility levels (50m, 75m, 150m) and KITTI-Rain with three precipitation rates (75mm, 100mm, 200mm). For Cityscapes validation images, we apply fog synthesis following Sakaridis et al.\cite{sakaridis2018model} with two severity parameters (0.01,0.02). Additionally, we evaluate on ACDC \cite{ACDC}, a real-world adverse weather dataset containing fog, rain, night, and snow conditions with the same object classes as Cityscapes. We adapt the Cityscapes-trained source model to each ACDC weather condition without accessing source data or target labels during adaptation.
\subsection{Implementation Details}
We implement our proposed TTA framework on Faster R-CNN with ResNet-50 backbone. Source model training details are provided in the supplementary material. The subtractive buffer is applied to all BN layers in bottleneck blocks, while additive buffers are inserted into residual shortcut paths. Source feature statistics are pre-computed off-line over the entire training set. 

Key hyperparameters are set as follows: Loss balance parameter $\lambda_{\text{reg}} = 0.05$,  suppression temperature $\lambda_S = 0.05$, compensation temperature $\lambda_A = 0.1$, target channel suppression ratio $5\%$, learning rate $\eta = 1 \times 10^{-4}$, and batch size $16$. We evaluate adaptation performance using mAP@50, the standard metric for object detection. To ensure statistical stability, all experimental results are reported as the mean of three independent runs with different random seeds.

\subsection{Baselines}
We compare our method against the following baselines representing different TTA paradigms:
\begin{itemize}

    \item \textbf{Direct Test} serves as the lower bound, evaluating the source model directly on target domains without adaptation to measure the performance gap caused by domain shift.
    \item \textbf{BufferTTA \cite{kim2025buffer}} introduces lightweight Buffer Layers as additive adapters, optimized through entropy minimization to encourage confident predictions on unlabeled target data.
    \item \textbf{PruningTTA \cite{wang2025pruning}} applies channel-sensitivity-based pruning to remove domain-sensitive channels, using KL divergence loss to align feature distributions between source and target domains.
    \item \textbf{ActMAD \cite{mirza2023actmad}} aligns feature statistics across all BN layers through multi-level distribution matching, minimizing L1 distance between target and source feature means and variances.
    \item \textbf{WHW \cite{yoo2024WHW}} inserts weighted adapter modules in parallel to backbone blocks, employing KL divergence loss to align both image-level and instance-level feature distributions for object detection.
\end{itemize}

\subsection{Main results}

\begin{table*} 
\centering
\small
\resizebox{\textwidth}{!}{%
\begin{tabular}{l|ccc|ccc|cc}
\hline
\multicolumn{1}{c|}{}&
  \multicolumn{3}{c|}{\textbf{KITTI fog}}&
  \multicolumn{3}{c|}{\textbf{KITTI rain}}&
  \multicolumn{2}{c}{\textbf{ Cityscapes foggy}}\\ \hline
\textbf{corruption/severity}&
  \textbf{fog/50m}&\textbf{fog/75m}&\textbf{fog/150m} &
  \textbf{rain/200mm}&\textbf{rain/100mm}& \textbf{rain/75mm}&
  \textbf{fog/0.02}&\textbf{fog/0.01}\\ \hline
\textbf{Direct Test}&
  21.27&30.84&50.45&
  47.11&65.32&69.03&
  11.87&18.38\\
\textbf{Buffer TTA \cite{kim2025buffer}}&
  23.21&33.12&52.50&
  50.96&69.28&\underline{73.43} &
  14.67&21.53\\
\textbf{Pruning TTA \cite{wang2025pruning}}&
  33.97&42.83&58.58&
  50.94&65.42&69.27&
  \underline{17.39}&20.90\\
\textbf{ActMAD \cite{mirza2023actmad}}&
  \underline{39.65}&\underline{49.95} & 60.18&
  56.37& 62.94& 64.63&
  16.53& \underline{21.74} \\
\textbf{WHW \cite{yoo2024WHW}}&
  34.38& 42.91& \underline{63.54}&
  \underline{61.17}&\textbf{73.15}& \textbf{74.33} &
  14.72&20.98\\
\textbf{Ours}&
  \textbf{44.80}&\textbf{56.06}&\textbf{68.42}&
  \textbf{63.22}&\underline{71.40}&73.03&
  \textbf{18.39}&\textbf{21.79}\\ \hline
\end{tabular}%
}
\vspace{2mm}
\caption{Quantitative results of object detection under various adverse weather conditions.
Performance (mAP@50) is reported for KITTI $\rightarrow$ KITTI (fog, rain) and Cityscapes $\rightarrow$ Cityscapes Foggy scenarios with different corruption severities.
The proposed method (Ours) achieves the best or comparably high performance in most cases, while other methods show inconsistent trends depending on severity.
In contrast, our method maintains robust and stable performance across diverse weather severities.}
\label{tab:table1}
\end{table*}


\begin{table}
\centering
\scriptsize  
\resizebox{\columnwidth}{!}{%
\begin{tabular}{l|cccc}
\hline
\multicolumn{1}{c|}{} & \multicolumn{4}{c}{\textbf{Cityscapes $\rightarrow$ ACDC}} \\\hline
\textbf{corruption} & \textbf{fog} & \textbf{snow} & \textbf{rain} & \textbf{night} \\ \hline
\textbf{Direct Test}  & 16.50  & 11.04 & 7.82  & 4.83 \\
\textbf{Buffer TTA \cite{kim2025buffer}}   & \underline{24.16} & \textbf{17.18} & 11.70  & \underline{6.98} \\
\textbf{Pruning TTA \cite{wang2025pruning}}  & 19.76 & 14.93 & 11.59 & 6.39 \\
\textbf{ActMAD \cite{mirza2023actmad}}       & 18.39 & 10.56 & 9.78  & 5.56 \\
\textbf{WHW \cite{yoo2024WHW}}          & 23.41 & \underline{16.99} & \underline{12.52} & 6.73 \\
\textbf{Ours}         & \textbf{24.45} & 15.41 & \textbf{13.71} & \textbf{8.92} \\ \hline
\end{tabular}%
}
\vspace{2mm}
\caption{Object detection results (mAP@50) for \textbf{Cityscapes $\rightarrow$ ACDC} across four corruption types (fog, snow, rain, night). 
The proposed method (Ours) achieves the best or comparably high performance with robust behavior across conditions.}
\label{tab:table2}
\end{table}

To comprehensively evaluate our framework, we conduct two types of experiments: \textbf{discrete TTA} with controlled severities to analyze performance across varying shift magnitudes, and \textbf{continual TTA} under progressive shifts to assess adaptation stability. All results are reported using mAP@50.
Table \ref{tab:table1}. presents results on KITTI fog/rain and Cityscapes fog, Table \ref{tab:table2}. shows ACDC real-world conditions, and Table \ref{tab:table3}. demonstrates continual adaptation under progressive domain shifts.
    \begin{figure}
    \centering
    \begin{subfigure}[t]{0.5\textwidth}
        \centering
        \includegraphics[width=\textwidth]{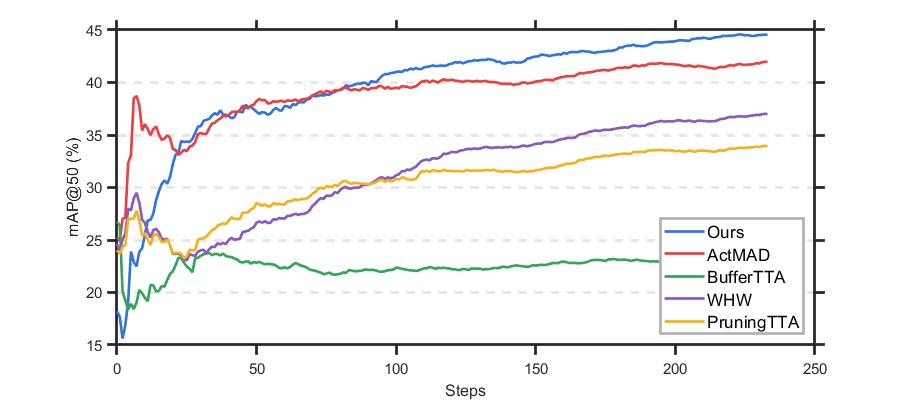}
        \caption{KITTI Fog 50m}
        \label{fig:fig3_a}
    \end{subfigure}%
    \hfill
    \begin{subfigure}[t]{0.5\textwidth}
        \centering
        \includegraphics[width=\textwidth]{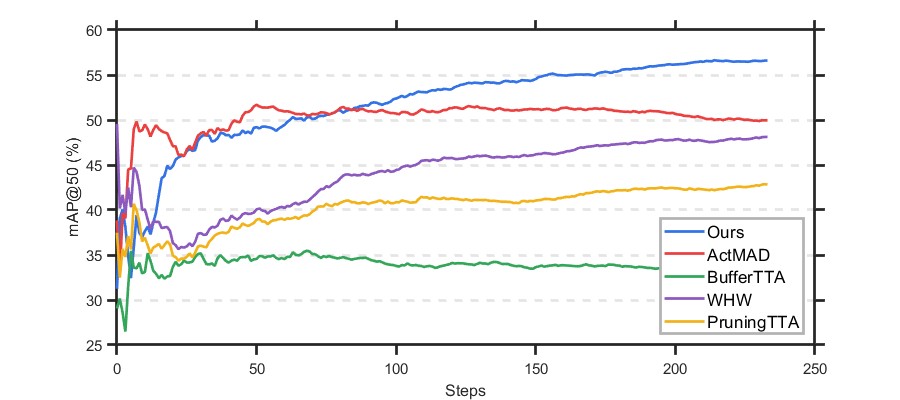}
        \caption{KITTI Fog 75m}
        \label{fig:fig3_b}
    \end{subfigure}
    \begin{subfigure}[t]{0.5\textwidth}
        \centering
        \includegraphics[width=\textwidth]{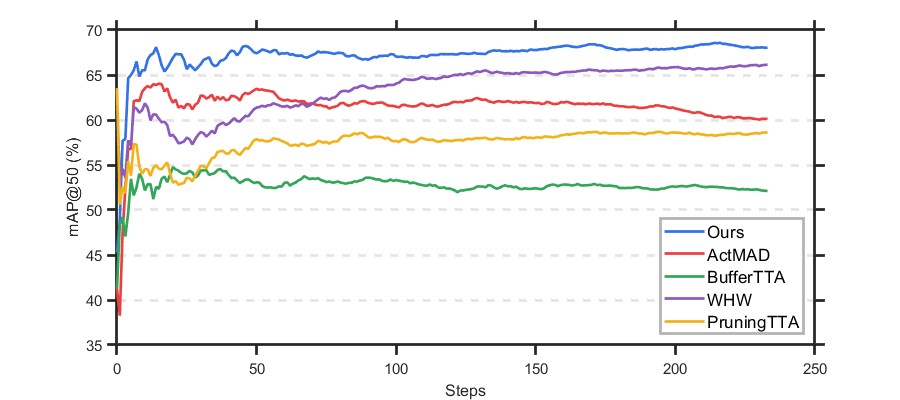}
        \caption{KITTI Fog 150m}
        \label{fig:fig3_c}
    \end{subfigure}

    \caption{Continual test-time adaptation (TTA) performance on KITTI Fog across different severities. The plots show the evolution of mAP@50 over adaptation steps for (a) 50m, (b) 75m, and (c) 150m fog distances. Our method (blue) achieves the fastest and most stable improvement, maintaining superior accuracy throughout adaptation.}
    \label{fig3}
    \end{figure}

\noindent{\textbf{Discrete TTA Results.}} Table \ref{tab:table1}. demonstrates our method's superior and consistent performance across varying domain shift severities. Existing methods exhibit inconsistent performance depending on weather type and severity. ActMAD shows comparable performance under severe conditions but experiences significant degradation under moderate corruption KITTI rain 100mm/75mm, Cityscapes fog 0.01. Similar inconsistency appears in Pruning TTA, Buffer TTA, and WHW. 

Notably, additive approaches BufferTTA, WHW achieve relatively strong performance under moderate shifts, particularly evident on KITTI rain scenarios, but suffer substantial degradation under severe conditions KITTI fog 50m/75m. Conversely, the subtractive approach like PruningTTA exhibits the opposite pattern, confirming the complementary characteristics of the two paradigms. This inconsistency is also observed in Table \ref{tab:table2}. on the ACDC dataset.

From a feature distribution perspective, severe domain shifts generate many features with large distances from source statistics. In such cases, subtractive methods prove effective by removing detrimental features and concentrating adaptation on robust channels. Under moderate shifts where feature distances are smaller, additive methods are more effective by refining features toward alignment. Our method addresses both scenarios through feature-level discrepancy-driven modulation: applying aggressive suppression to severely shifted features while providing refinement via additive buffers for moderately shifted features. Simultaneously, suppressed channels receive strong compensation from the additive buffer, whereas the remaining channels are softly scaled according to their distance from source statistics. This mechanism enables consistently strong performance across diverse severities and conditions.

Figure \ref{fig3}. illustrates performance trajectory across adaptation steps on KITTI fog with three severity levels. Our method demonstrates stable improvement trends with rapid adaptation, while ActMAD shows sharp initial gains but unstable convergence. The comparison again reveals inconsistent adaptation capabilities of baseline methods depending on severity, whereas our method maintains consistency throughout the adaptation process as well as in final performance. In particular, Figure \ref{fig:fig3_c}. shows that our method not only adapts to the target domain most rapidly but also sustains the highest performance throughout adaptation, highlighting its superior consistency and robustness.


\begin{table*}[!ht]
\centering
\resizebox{\textwidth}{!}{%
\begin{tabular}{l|ccc|ccc|ccc|ccc|c} 
\hline
& \multicolumn{12}{c}{\textbf{KITTI $\rightarrow$ KITTI fog (50m $\rightarrow$ 75m $\rightarrow$ 150m)}} & \\ \hline

\multicolumn{1}{l|}{} &
  \multicolumn{3}{c}{\textbf{Round1}} &
  \multicolumn{3}{c}{\textbf{Round2}} &
  \multicolumn{3}{c}{\textbf{Round5}} &
  \multicolumn{3}{c|}{\textbf{Round10}} &
  \multirow{2}{*}{\textbf{Average}} \\
\cline{2-13}
\multicolumn{1}{l|}{} &
  \textbf{50m} & \textbf{75m} & \textbf{150m} &
  \textbf{50m} & \textbf{75m} & \textbf{150m} &
  \textbf{50m} & \textbf{75m} & \textbf{150m} &
  \textbf{50m} & \textbf{75m} & \textbf{150m} & \\ \hline

\multicolumn{1}{l|}{\textbf{Direct Test}} &
  21.27 & 30.84 & 50.45 &
  21.27 & 30.84 & 50.45 &
  21.27 & 30.84 & 50.45 &
  21.27 & 30.84 & 50.45 &
  34.19 \\

\multicolumn{1}{l|}{\textbf{Buffer TTA \cite{kim2025buffer}}} &
  22.85 & 32.48 & 52.11 &
  23.56 & 32.45 & 52.24 &
  24.15 & 34.52 & 54.06 &
  27.65 & 38.93 & \underline{58.85} &
  37.82 \\

\multicolumn{1}{l|}{\textbf{Pruning TTA \cite{wang2025pruning}}} &
  24.45 & 39.45 & 60.15 &
  30.09 & 43.78 & 62.02 &
  33.96 & 44.97 & 62.05 &
  32.60 & 43.07 & 57.79 &
  44.53 \\

\multicolumn{1}{l|}{\textbf{ActMAD \cite{mirza2023actmad}}} &
  \textbf{42.07} & \textbf{54.55} & 64.65 &
  \underline{42.21} & \underline{52.31} & 61.03 &
  38.87 & 47.71 & 54.11 &
  34.22 & 41.41 & 46.66 &
  48.32 \\

\multicolumn{1}{l|}{\textbf{WHW \cite{yoo2024WHW}}} &
  23.96 & 39.14 & \underline{65.51} &
  39.21 & 51.53 & \underline{68.43} &
  \underline{42.38} & \underline{52.18} & \textbf{68.32} &
  \underline{43.74} & \underline{52.68} & 54.96 &
  \underline{50.17} \\

\multicolumn{1}{l|}{\textbf{Ours}} &
  \underline{34.90} & \underline{52.70} & \textbf{69.93} &
  \textbf{44.51} & \textbf{57.29} & \textbf{70.72} &
  \textbf{46.53} & \textbf{58.30} & \underline{68.02} &
  \textbf{45.40} & \textbf{56.14} & \textbf{63.51} &
  \textbf{55.66} \\ \hline
\end{tabular}%
}
\vspace{2mm}
\caption{Continual test-time adaptation (TTA) results for the KITTI $\rightarrow$ KITTI fog scenario.
This experiment simulates progressive domain shift where fog severity gradually increases (50\,m $\rightarrow$ 75\,m $\rightarrow$ 150\,m). Each model continuously adapts over 10 rounds, using outputs from one severity level as initialization for the next. The results demonstrate the robustness of each method under incremental domain changes, with \textit{Ours} consistently achieving superior or stable performance across all adaptation rounds and the highest overall average mAP@50.}
\label{tab:table3}
\end{table*}

\noindent\textbf{Continual TTA Results.} Table \ref{tab:table3}. presents the continual TTA results on KITTI Fog across three severity levels, where models adapt over 10 rounds by using the output of each severity as initialization for the next. Our method consistently achieves the best or second-best performance across all rounds and severity transitions, yielding the highest overall mAP@50.

As shown in Figure \ref{fig3}., ActMAD exhibits rapid initial adaptation but suffers from instability during extended adaptation cycles. This instability reflects the inherent limitations of full-parameter update strategies, which tend to accumulate errors and induce catastrophic forgetting under continuously shifting domains \cite{tian2024parameter}. In contrast, other methods achieve slower yet steadier convergence, although often with limited capacity to adapt to more severe shifts.

Our method demonstrates both rapid early-stage adaptation and consistently improving performance across rounds. The complementary dual-buffer mechanism enables this behavior by simultaneously suppressing out-of-distribution features and applying scaled refinement to channels requiring adjustment, producing strong initial performance and stable transitions across domain shifts.

\begin{table}
\centering
\large
\resizebox{\columnwidth}{!}{%
\begin{tabular}{lcccc|c}
\toprule
\textbf{$\mathcal{L}_{\text{mask}}$} & \textbf{BN layer} & \textbf{Subtractive} & \textbf{Additive} & \textbf{Grad scaling} & \textbf{mAP@50} \\
\midrule
-- & \checkmark & -- & -- & -- & 8.69 \\
-- & \checkmark & \checkmark & -- & -- & 15.69 \\
\checkmark & \checkmark & \checkmark & -- & -- & 16.81 \\
\checkmark & \checkmark & -- & \checkmark & -- & 18.66 \\
\checkmark & \checkmark & \checkmark & \checkmark & -- & 18.74 \\
-- & \checkmark & \checkmark & \checkmark & \checkmark & 18.97 \\
\checkmark & \checkmark & \checkmark & \checkmark & \checkmark & 19.13 \\
\bottomrule
\end{tabular}%
}
\caption{Ablation study evaluating the contribution of each component in our framework. A checkmark (\checkmark) indicates that the corresponding component is enabled. All experiments are performed on the Cityscapes Fog 0.02 dataset.}
\label{tab:table4}
\end{table}

Furthermore, unlike Pruning TTA, which directly regularizes BN affine parameters $\gamma$ for channel suppression and risks irreversible information loss, our method introduces independent learnable mask scores for both suppression and compensation, thereby minimizing degradation of source knowledge. Moreover, while adapter-based approaches rely solely on learnable parameters for adaptation, our framework explicitly guides updates based on the measured discrepancy between source and target features. This principled guidance substantially enhances adaptation efficiency and robustness, enabling reliable continual TTA under diverse and progressively evolving domain shifts.

\subsection{Ablation study}
Table \ref{tab:table4}. resents an ablation study evaluating the contribution of each component in our framework. We analyze the impact of the discrepancy-driven masking loss $\mathcal{L}_{\text{mask}}$, subtractive buffer, additive buffer, and layer-wise scaling.

\noindent\textbf{Importance of Discrepancy-Driven Guidance.} The most critical finding concerns the role of $\mathcal{L}_{\text{mask}}$. Adapting only BN layers achieves 8.69 mAP@50, whereas introducing the subtractive buffer with $\mathcal{L}_{\text{mask}}$ significantly improves performance to 16.81 (+8.12). However, removing $\mathcal{L}_{\text{mask}}$ while keeping the subtractive buffer causes performance to drop to 15.69, as the mask scores are updated solely through the alignment loss without explicit discrepancy guidance. This comparison validates the critical role of discrepancy-weighted regularization: without it, the framework lacks explicit guidance on which channels to suppress based on domain shift severity. The $\mathcal{L}_{\text{mask}}$ term encodes our core principle that channels with higher feature-level discrepancy $D_c$ receive stronger regularization, providing explicit guidance for both suppression and compensation. The additive buffer alone surpasses the subtractive buffer by +1.85 mAP@50, indicating that feature refinement provides stronger baseline adaptation than suppression under severe domain shift.

\begin{figure}[t]
    \centering
    \begin{subfigure}[t]{0.48\linewidth}
        \centering
        \includegraphics[width=\linewidth]{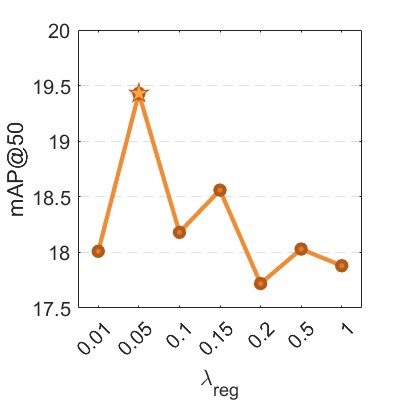}
        \caption{Effect of $\lambda_{\mathrm{reg}}$}
        \label{fig:lambda}
    \end{subfigure}%
    \hfill
    \begin{subfigure}[t]{0.48\linewidth}
        \centering
        \includegraphics[width=\linewidth]{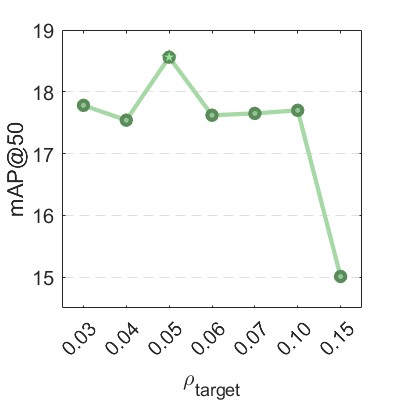}
        \caption{Effect of $\rho_\text{target}$}
        \label{fig:ratio}
    \end{subfigure}

    \caption{Effect of key hyperparameters on TTA performance:(a) regularization weight $\lambda_{\mathrm{reg}}$ and (b) pruning ratio $\rho_{\text{target}}$.}
    \label{fig:tta_params}
\end{figure}

Figure \ref{fig:tta_params}. analyzes the sensitivity of our framework to key hyperparameters: regularization weight $\lambda_{\text{reg}}$ and channel suppression ratio $\rho_t$. The regularization weight exhibits a clear optimal range around 0.05, where insufficient regularization (0.01) fails to suppress corrupted channels while excessive regularization (0.5) causes severe information loss. Similarly, the suppression ratio shows stable performance within 2-3\%, with degradation occurring when too few channels are suppressed or too many are removed. We selected the hyperparameters in a balanced manner based on these experimental observations. 
\section{Conclusion}
\label{sec:Conclusion}
We presented CD-Buffer, a complementary dual-buffer framework for test-time adaptation in adverse weather object detection. We revealed that existing additive and subtractive TTA methods exhibit complementary failure patterns depending on domain shift severity, motivating our unified approach that dynamically balances both strategies based on feature-level domain discrepancy.

Our framework introduces a subtractive buffer with learnable mask scores for channel suppression and an additive buffer providing inverse-modulated compensation, both unified by discrepancy-driven guidance. Extensive experiments demonstrate state-of-the-art performance across KITTI, Cityscapes, and ACDC datasets, consistently outperforming existing methods across diverse weather conditions and severity levels. Ablation studies validate that discrepancy-driven coupling is essential for synergistic dual-buffer adaptation.

\clearpage

\noindent{\textbf{Acknowledgements.}} This work was supported by the National Research Foundation of Korea (NRF) grant funded by the Korea government (MSIT) (No. RS-2025- 16070382, RS-2025-02215070, RS-2025-02217919), Artificial Intelligence Graduate School Program at Yonsei University (RS-2020-II201361), the Korea Institute of Science and Technology (KIST) Institutional Program under Grant 26E0170.
{
    \small
    \bibliographystyle{ieeenat_fullname}
    \bibliography{main}
}

\clearpage
\clearpage
\renewcommand*{\thesection}{\Alph{section}}
\setcounter{page}{1}
\maketitlesupplementary
\renewcommand{\theequation}{A.\arabic{equation}}
\setcounter{equation}{0}

\setcounter{section}{0}
\section{Overview}
This supplementary material offers comprehensive details and additional experimental evidence supporting our main contributions. Section \ref{secB} describes implementation specifics, covering source model training, baseline method configurations, and additional CD-Buffer details. Section \ref{secC} analyzes CD-Buffer's design choices and introduces CD-Buffer Light, an efficient variant optimized for real-time deployment, along with its performance evaluation across various adaptation scenarios. Section \ref{secD} presents supplementary experimental results, including visualizations of the continual test-time adaptation process and qualitative detection results under diverse weather conditions.

\section{Additional Implementation Details}
\label{secB}
\textbf{Hardware:} All experiments, including source model training and test-time adaptation, were performed using a single NVIDIA RTX 6000 Ada Generation GPU.

\noindent\textbf{Source model training:} For TTA experiments, we trained source model on the KITTI and Cityscapes datasets using the following configurations. All source models adopt the Faster R-CNN detector with a ResNet-50~\cite{he2016deep} backbone initialized from ImageNet-pretrained weights~\cite{deng2009imagenet}. We trained the models using the Adam optimizer with a batch size of 32 for a total of 250 epochs. The learning rates were set to 1e-4 for KITTI and 5e-4 for Cityscapes, respectively. The datasets contain eight object categories each:
\begin{itemize}
    \item \textbf{KITTI:} Car, Van, Truck, Pedestrian, Person sitting, Cyclist, Tram, Misc.
    \item \textbf{Cityscapes:} Person, Rider, Car, Truck, Bus, Train, Motorcycle, Bicycle.
\end{itemize}
The resulting source model performance, measured by mAP@50, is 85.75 on KITTI, and 24.75 on Cityscapes.

\noindent\textbf{Baselines implementation:} We provide detailed implementation settings for all baseline methods. For fair comparison, we unified the batch size to 16 across all experiments and selected learning rates by testing both the originally reported values and the learning rate used in our method, choosing whichever yielded better performance.

\noindent\textbf{BufferTTA.} Following the original paper's best-performing configuration, we introduce parallel $1\times1$ and $3\times3$ convolutional layers as buffer layers. These buffer layers are inserted after ReLU activation functions. Among four stages of ResNet-50 backbone, we experimented with buffer placement in each stage and selected Stage 3, which achieved the highest performance. The learning rate is set to $1 \times 10^{-5}$ uniformly across all datasets. Learnable parameters include batch normalization affine parameters and the introduced buffer parameters.

\begin{algorithm}[htbp]
\caption{CD-Buffer Test-Time Adaptation}
\label{alg:cdbuffer}
\begin{algorithmic}[1]
\REQUIRE Source domain statistics $\{\bar{X}_s^l, \bar{x}_s^l, \bar{\mu}_s^l, \bar{\sigma}_s^l\}$, $\lambda_S, \lambda_A, \lambda_{\text{reg}}, \rho_{\text{target}}$, $r$.

\STATE Initialize mask scores $s = \lvert \gamma \rvert$.
\FOR{each target batch $\mathcal{B}_t$ in the test stream}
    \FOR{each adaptable block}
        \STATE Let $F_{\text{in}}$ denote the input feature of the adaptable block.
        \FOR{each layer $l$}
            \STATE Let $X_t^{l}$ denote the input feature of layer $l$.
            \STATE Compute $m_{\text{hard}}$ and $m$ for the subtractive buffer as in Eqs.~(6)--(8).
            \STATE \textbf{Step1: Subtractive buffer}:
            \STATE Compute $F^{s}_{\text{out}}$ as in Eq.~(4).
        \ENDFOR
        \STATE Compute $\hat{m}^{-1}_{\text{soft}}$ for the additive buffer as in Eqs.~(10) and~(11).
        \STATE \textbf{Step2: Additive buffer}:
        \STATE Compute $F^{\text{a}}_{\text{out}}$ as in Eqs.~(9) and~(12).
        \STATE $F_{\text{out}} = F^{s}_{\text{out}} + F^{\text{a}}_{\text{out}}$.
    \ENDFOR
    \STATE \textbf{Step3: Compute losses}
    \STATE Compute domain discrepancy score $D$ as in Eqs.~(1)--(3).
    \STATE Compute total loss $\mathcal{L}_{\text{total}}$ as in Eqs.~(5), (13), and~(14).

    \STATE \textbf{Step4: Optimization}
    \STATE Compute $D_{l}$ as in Eq.~(15).
    \STATE Update mask scores $s^{l}$ and BN layer parameters.
    \STATE Update additive buffer parameters scaled by $D_{l}$.

    \STATE \textbf{Step5: Stochastic reactivation}
    \FOR{each suppressed channel with $m^{c,l} = 0$}
        \STATE Sample a random variable $v_c^{l}$ with 
        \STATE $v_c^{l} \sim \text{Bernoulli}(r)$.
        \IF{$v_c^{l} = 1$}
            \STATE $s_c^{l} = \lvert \gamma_c^{l} \rvert$.
        \ENDIF
    \ENDFOR
\ENDFOR
\end{algorithmic}
\end{algorithm}

\noindent\textbf{Pruning TTA.} Since official code is not publicly available, we implement this method following the paper's description. The approach is applied to all BN layers in the ResNet-50 backbone. For fair comparison, we set the pruning ratio threshold to 0.05, matching our suppression ratio. The learning rate is set to $1 \times 10^{-4}$. All other hyperparameters follow the reported values in the paper. Learnable parameters consist of scaling factors in BN layers while freezing all other parameters. The loss function employs KL divergence between source and target feature distributions for all adaptable BN layer inputs, using both image-level and instance-level features.

\noindent\textbf{ActMAD.} Apply L1 loss between source and target batch statistics (mean and variance) for all BN layer output features. Unlike other methods, all model parameters are updated during adaptation. The learning rate is set to $1 \times 10^{-4}$. All other hyperparameter settings follow the configurations reported in the original paper.

\noindent\textbf{WHW.} This method aligns source and target feature distributions using KL divergence loss, utilizing both image-level and instance-level features. For stable statistical estimation of target feature statistics, we employ exponential moving average (EMA). The learning rate is set to $1 \times 10^{-4}$ for all datasets. All other hyperparameters follow the reported settings.

\noindent\textbf{Additional CD-Buffer Details:} CD-Buffer is applied to all basic blocks across all stages of the ResNet-50 backbone network. Within each block, the subtractive buffer is applied to every BN layer, while one additive buffer is applied per block. The scaling factor $k$ for determining the range of $\hat{m}^{-1}_{\text{soft}}$ is set to 10, resulting in adaptive channel-wise scaling within the range $[0, 10]$ for the additive buffer. To prevent early suppression, we introduce a stochastic reactivation mechanism following the approach in prior work~\cite{wang2025pruning}. At each adaptation step, suppressed channels (where $m_c = 0$) are reactivated with probability $r$ through Bernoulli sampling:
\begin{equation}
v_c \sim \text{Bernoulli}(r), \quad \text{if } v_c = 1: s_c \leftarrow |\gamma_c|,
\end{equation}
where $v_c$ is a random variable determining reactivation, and the mask score $s_c$ is reset to the absolute value of the corresponding BN parameter $\gamma_c$. The reactivation probability is set to $r = 0.05$ across all experiments. The complete TTA process is presented in Algorithm \ref{alg:cdbuffer}.

\begin{table*} 
\centering
\small
\resizebox{\textwidth}{!}{%
\begin{tabular}{l|ccc|ccc|cc}
\hline
\multicolumn{1}{c|}{}&
  \multicolumn{3}{c|}{\textbf{KITTI fog}}&
  \multicolumn{3}{c|}{\textbf{KITTI rain}}&
  \multicolumn{2}{c}{\textbf{ Cityscapes foggy}}\\ \hline
\textbf{corruption/severity}&
  \textbf{fog/50m}&\textbf{fog/75m}&\textbf{fog/150m} &
  \textbf{rain/200mm}&\textbf{rain/100mm}& \textbf{rain/75mm}&
  \textbf{fog/0.02}&\textbf{fog/0.01}\\ \hline
\textbf{Ours}&
  \textbf{44.80}&\textbf{56.06}&\textbf{68.42}&
  \textbf{63.22}&\underline{71.40}&73.03&
  \textbf{18.39}&\textbf{21.79}\\ \hline
\textbf{Stage 4}&
  36.91&44.69&58.79&
  53.65&65.11&67.19&
  12.50&17.25\\ 
\textbf{Stage 3}&
  \textbf{42.19}&\textbf{50.89}&\textbf{65.64}&
  \textbf{62.21}&\underline{70.70}&72.05&
  \textbf{17.48}&\textbf{22.51}\\ 
\textbf{Stage 2}&
  \textbf{43.28}&\textbf{53.99}&\textbf{67.26}&
  \textbf{63.09}&\underline{71.39}&72.58&
  \textbf{17.94}&20.63\\ 
\textbf{Stage 1(Ours-Light)}&
  \textbf{42.51}&\textbf{52.04}&\textbf{66.95}&
  \textbf{61.85}&\underline{70.88}&73.08&
  \textbf{18.57}&\textbf{22.49}\\ \hline
\end{tabular}%
}
\vspace{2mm}
\caption{Quantitative results of object detection under various adverse weather conditions.
Performance (mAP@50) is reported for KITTI $\rightarrow$ KITTI (fog, rain) and Cityscapes $\rightarrow$ Cityscapes Foggy scenarios with different corruption severities. We highlight entries that still achieve the best performance compared to the baselines. in bold, and those with the second-best performance with an underline. We propose Stage 1 as Ours-Light, which consistently maintains strong performance across various corruptions and severity levels.}
\label{tab:table5}
\end{table*}
\begin{table*}[!ht]
\centering
\resizebox{\textwidth}{!}{%
\begin{tabular}{l|ccc|ccc|ccc|ccc|c} 
\hline
& \multicolumn{12}{c}{\textbf{KITTI $\rightarrow$ KITTI fog (50m $\rightarrow$ 75m $\rightarrow$ 150m)}} & \\ \hline

\multicolumn{1}{l|}{} &
  \multicolumn{3}{c}{\textbf{Round1}} &
  \multicolumn{3}{c}{\textbf{Round2}} &
  \multicolumn{3}{c}{\textbf{Round5}} &
  \multicolumn{3}{c|}{\textbf{Round10}} &
  \multirow{2}{*}{\textbf{Average}} \\
\cline{2-13}
\multicolumn{1}{l|}{} &
  \textbf{50m} & \textbf{75m} & \textbf{150m} &
  \textbf{50m} & \textbf{75m} & \textbf{150m} &
  \textbf{50m} & \textbf{75m} & \textbf{150m} &
  \textbf{50m} & \textbf{75m} & \textbf{150m} & \\ \hline

\multicolumn{1}{l|}{\textbf{Ours}} &
  \underline{34.90} & \underline{52.70} & \textbf{69.93} &
  \textbf{44.51} & \textbf{57.29} & \textbf{70.72} &
  \textbf{46.53} & \textbf{58.30} & \underline{68.02} &
  \textbf{45.40} & \textbf{56.14} & \textbf{63.51} &
  \textbf{55.66} \\
  
  \multicolumn{1}{l|}{\textbf{Ours-Light}} &
  \underline{30.18} & \underline{49.10} & \textbf{67.92} &
  \underline{40.36} & \textbf{53.52} & \textbf{69.14} &
  \textbf{45.52} & \textbf{58.02} & \textbf{70.24} &
  \textbf{47.20} & \textbf{59.54} & \textbf{68.49} &
  \textbf{54.94} \\ \hline
\end{tabular}%
}
\vspace{2mm}
\caption{Continual test-time adaptation (TTA) results for the KITTI $\rightarrow$ KITTI fog scenario.
This experiment simulates progressive domain shift where fog severity gradually increases (50\,m $\rightarrow$ 75\,m $\rightarrow$ 150\,m). Each model continuously adapts over 10 rounds, using outputs from one severity level as initialization for the next. We highlight entries that still achieve the best performance compared to the baselines. in bold, and those with the second-best performance with an underline.}
\label{tab:table7}
\end{table*}
\begin{table}
\centering
\tiny  
\resizebox{\columnwidth}{!}{%
\begin{tabular}{l|cccc}
\hline
\multicolumn{1}{c|}{} & \multicolumn{4}{c}{\textbf{Cityscapes $\rightarrow$ ACDC}} \\\hline
\textbf{corruption} & \textbf{fog} & \textbf{snow} & \textbf{rain} & \textbf{night} \\ \hline
\textbf{Ours}         & \textbf{24.45} & 15.41 & \textbf{13.71} & \textbf{8.92} \\
\textbf{Ours-Light}         & 24.15 & 16.97 & \textbf{15.15} & \textbf{8.18} \\ \hline
\end{tabular}%
}
\vspace{2mm}
\caption{Object detection results (mAP@50) for \textbf{Cityscapes $\rightarrow$ ACDC} across four corruption types (fog, snow, rain, night). We highlight entries that still achieve the best performance compared to the baselines in Table. in bold, and those with the second-best performance with an underline.}
\label{tab:table6}
\end{table}

\begin{table}
\centering
\tiny
\resizebox{0.5\columnwidth}{!}{%
\begin{tabular}{l|cc}
\hline
                     & \textbf{FPS}   \\ \hline
\textbf{Direct Test} & 73.24          \\
\textbf{Buffer TTA}  & 33.69          \\
\textbf{Pruning TTA} & \textbf{40.62} \\
\textbf{ActMAD}      & 35.59          \\
\textbf{WHW}         & 31.38          \\
\textbf{Ours}        & 15.59          \\
\textbf{Ours\_Light} & \underline{35.86}    \\ \hline
\end{tabular}%
}
\caption{Inference speed comparison (FPS) for different TTA methods measured on a dummy input. The highest FPS is highlighted in bold, and the second-highest is underlined.}
\label{tab:table8}
\end{table}

\section{CD-buffer Light: Efficient Variant for Real-Time Applications}
\label{secC}
Given the real-time nature of object detection tasks, processing speed (FPS) is as critical as adaptation performance. While applying CD-Buffer to all stages of the backbone network achieves optimal performance, it may introduce computational overhead unsuitable for time-critical applications. Therefore, we conduct an ablation study on CD-Buffer placement across different stages and propose CD-Buffer Light, a more efficient variant tailored for real-world scenarios.
    \begin{figure*}
        \centering
        \includegraphics[width=\textwidth]{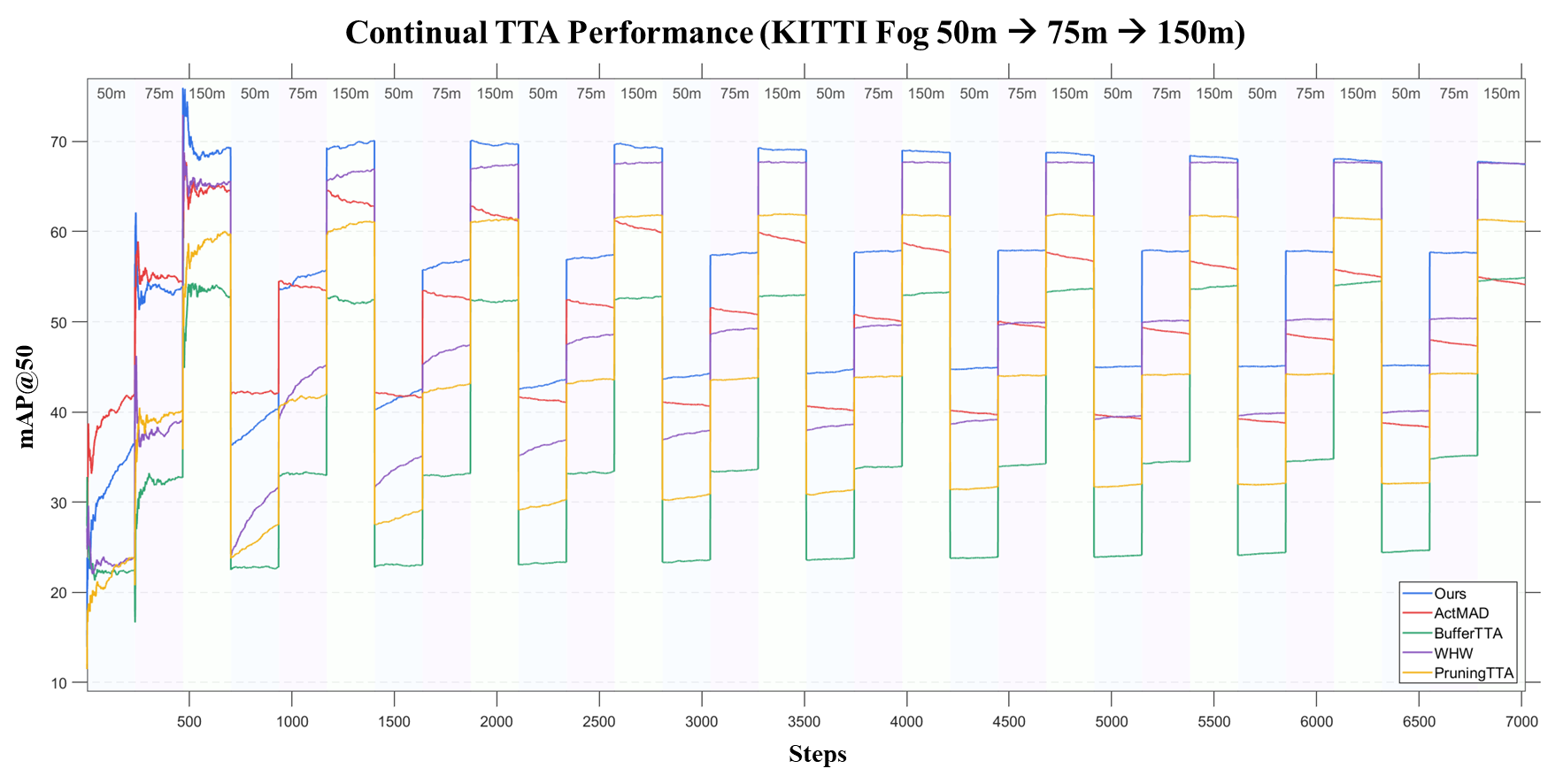}
        \caption{\textbf{Visualization of continual TTA process on KITTI fog with cyclic severity transitions} (50m $\rightarrow$ 75m $\rightarrow$ 150m, repeated 10 times). Our method maintains consistently superior performance across all cycles, demonstrating both rapid adaptation within each severity level and robustness to various severity.}
        \label{fig5.CTTA_KITTI_plot}
    \end{figure*}

\subsection{CD-buffer Placement Analysis}
Table \ref{tab:table5}. presents TTA results on KITTI and Cityscapes datasets, comparing our fully adaptation model (applied to all stages) against variants where CD-Buffer is applied to only one stage (Stage 1, 2, 3, or 4) of the ResNet-50 backbone. Single-stage variants exhibit performance degradation compared to the fully adaptation model, but most configurations still outperform baseline methods. Notably, Stage 4 shows the weakest performance among single-stage variants. In contrast, Stage 1 demonstrates the most consistent performance across all severity levels, achieving the best average performance among single-stage variants. Moreover, comparing with baseline methods, Stage 1 outperforms most baselines across nearly all conditions. This demonstrates that our framework's performance does not degrade linearly with the number of learnable parameters, maintaining competitive results even with significantly reduced parameters. Based on these observations, we propose \textbf{CD-Buffer Light}, which applies our framework exclusively to Stage 1, as it provides the optimal balance between adaptation effectiveness and computational efficiency.
\subsection{Performance Evaluation of CD-Buffer Light}

Tables \ref{tab:table7} and \ref{tab:table6} present additional experimental results for CD-Buffer Light on continual adaptation scenarios and ACDC, respectively. Across both settings, CD-Buffer Light consistently outperforms baseline methods, validating its effectiveness despite reduced computational cost. While performance is slightly lower than the full model, the gap is modest.
\subsection{Inference Speed Comparison}
Table \ref{tab:table8} compares inference speed in FPS for all baseline methods, our fully adaptation model, and Ours-Light. The full CD-Buffer model is slower than most baselines due to buffers applied across all basic blocks. In contrast, CD-Buffer Light achieves significantly faster inference, outperforming all methods except Pruning TTA while maintaining superior adaptation performance.

Our framework offers flexibility for different deployment scenarios. CD-Buffer (Full) is recommended when accuracy is prioritized, while CD-Buffer Light suits real-time applications with strict latency constraints.

\section{Additional Results}
\label{secD}
\noindent\textbf{Continual TTA visualization on KITTI dataset}

Figure \ref{fig5.CTTA_KITTI_plot} visualizes the adaptation dynamics under cyclic domain shifts, where fog severity repeatedly transitions between 50m, 75m, and 150m over 10 rounds on the KITTI dataset. Our method maintains consistently superior performance throughout all cycles, demonstrating stable and robust adaptation across repeated domain transitions.

\noindent\textbf{Qualitative results}
Figures \ref{fig6.kitti_fog_qual}, \ref{fig7}, and \ref{fig8} present qualitative visualization of object detection results across diverse adverse weather conditions. We compare three settings: Direct Test (source model without adaptation), Ground Truth annotations, and our CD-Buffer adaptation results. The visualizations cover KITTI fog and rain scenarios at various severity levels, Cityscapes fog conditions, and ACDC including fog, snow, rain, and night conditions. The qualitative results clearly demonstrate that our method effectively addresses domain shift challenges at test time.

\begin{table}[hbt!]
\centering
\resizebox{\columnwidth}{!}{%
\begin{tabular}{
c|
>{\centering\arraybackslash}m{2.3cm}
>{\centering\arraybackslash}m{2.5cm}
>{\centering\arraybackslash}m{3.0cm}
>{\centering\arraybackslash}m{3.1cm}
}
\hline
\textbf{Fog} &
\textbf{Buffer TTA (Add)} &
\textbf{Pruning TTA (Sub)} &
\textbf{Parallel} &
\textbf{Ours} \\
\hline
\textbf{30m (Severe)} &
\hlb{14.54} &
\hlo{23.89} &
18.58 {\textbf{(\textcolor{blue}{\hlb{+4.04}}/ \textcolor{red!70!black}{\hlo{-5.31}})}} &
\textbf{33.04} {\textbf{(\textcolor{blue}{\hlb{+18.50}}/ \textcolor{blue}{\hlo{+9.15}})}} \\

\textbf{375m (Mild)} &

\hlb{73.70} &
\hlo{70.60} &
72.11 {\textbf{(\textcolor{red!70!black}{\hlb{-1.59}}/ \textcolor{blue}{\hlo{+1.51}})}} &
\textbf{74.97} {\textbf{(\textcolor{blue}{\hlb{+1.27}}/ \textcolor{blue}{\hlo{+4.37}})}} \\
\hline
\end{tabular}%
}
\caption{Parallel combination vs CD-Buffer on KITTI.}
\label{tab:table_re_1}
\end{table}

\noindent\textbf{Analysis of Complementary Behaviors}.

Our contribution lies not in the additive and subtractive components themselves, but in how they are coupled through a unified discrepancy metric. Prior methods treat them as independent alternatives, but we reformulate this as a channel-wise control problem. In Tab.~\ref{tab:table_re_1}, additive and subtractive methods exhibit opposite performance trends depending on severity, and their parallel combination fails to resolve this trade-off and can even degrade performance in certain cases (\textcolor{red}{\textbf{red}}). CD-Buffer achieves the best performance under both severe and mild conditions. The inverse soft mask enables automatic balancing, where suppression extent directly modulates compensation strength. In practical TTA where method selection cannot be made at runtime, consistent performance across diverse severities is a meaningful contribution.

\begin{table}[hbt!]
\centering
\resizebox{\columnwidth}{!}{%
\begin{tabular}{ccccc}
\hline
 & \textbf{30m} & \textbf{50m} & \textbf{75m} & \textbf{150m} \\ \hline
\textbf{w/o Sub}      & \textbf{31.88(-1.18)} & \textbf{44.51(-1.15)} & 55.67(-0.81) & 67.88(-0.45) \\
\textbf{w/o Add}      & 32.96(-0.10) & 45.22(-0.44) & \textbf{55.18(-1.30)} & \textbf{65.75(-2.58)} \\
\textbf{Full}  & 33.06 & 45.66 & 56.48 & 68.33 \\ \hline
\end{tabular}%
}

\caption{Severity-wise ablation on KITTI fog.}

\label{tab:table_re_3}
\end{table}
Complementarity is further characterized by severity-dependent roles rather than uniform contribution. In Tab.~\ref{tab:table_re_3}, the subtractive module contributes more under severe shifts (30m, 50m), while the additive module is more effective under milder shifts (75m, 150m), and the full model achieves the most consistent performance across conditions.
The modest gain (+0.08) observed in main paper can be attributed to the use of Cityscapes-Fog, which predominantly exhibits mild shifts. Under such conditions, the additive module naturally dominates, consistent with our analysis. The primary benefit of the proposed approach lies in stable adaptation across diverse shift severities rather than large average gains.

\begin{table}[hbt!]
\centering

\resizebox{0.75\columnwidth}{!}{%
\tiny
\begin{tabular}{cccc}
\hline
               & \textbf{L1}    & \textbf{L2} & \textbf{Cosine} \\ \hline
\textbf{mAP50} & \textbf{63.61} & 63.17       & 63.50                       \\ \hline
\end{tabular}%
}
\caption{Comparison of discrepancy metrics.}
\label{table_re_2}
\end{table}

\begin{figure}[hbt!]
    \centering
    \includegraphics[width=\columnwidth]{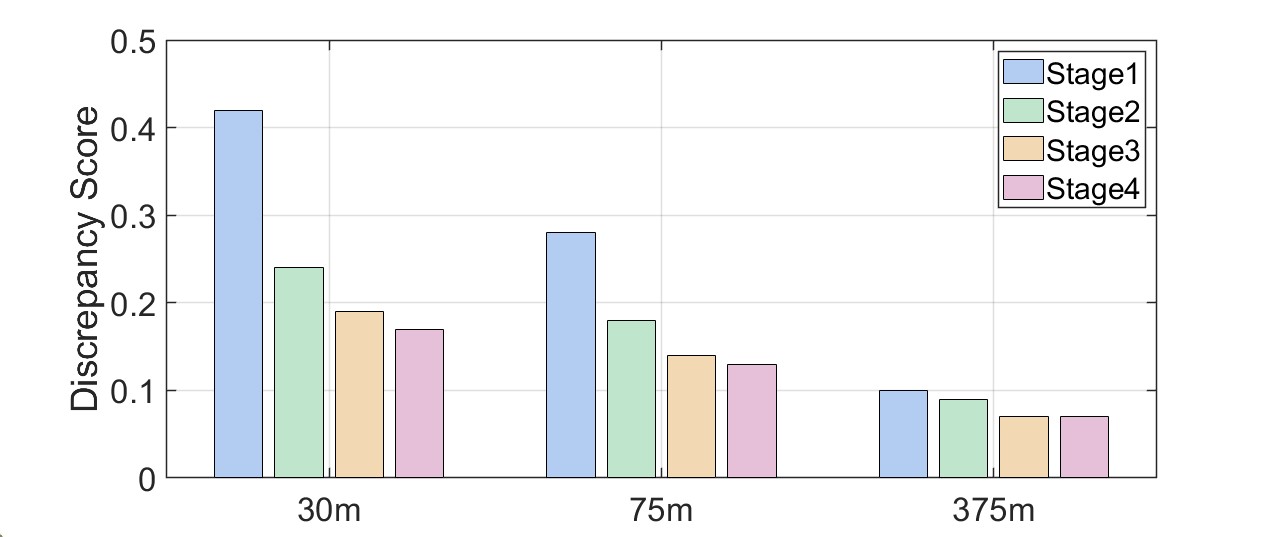}
    \caption{Discrepancy across fog severities and backbone stages.}
    \label{fig_re_1}
\end{figure}

\noindent{\textbf{Validation of the Discrepancy Metric}}.

To justify the choice of the discrepancy metric, we adopt L1 as a simple and effective measure of shift magnitude, as supported by prior TTA methods (e.g., ActMAD, PruningTTA). In Fig.~\ref{fig_re_1}, the discrepancy consistently decreases as visibility increases, confirming that the metric reliably reflects domain shift. A stage-wise analysis further shows that Stage 1 exhibits the highest discrepancy; applying CD-Buffer only to Stage 1 achieves the best single-stage performance Tab. ~\ref{tab:table5}, indicating that the metric also captures location-specific shifts. In Tab.~\ref{table_re_2}, we compare L1 with L2 and cosine similarity. While the differences are marginal, L1 achieves the best performance, supporting it as a simple yet effective and empirically validated choice for our framework.


\begin{table}[hbt!]
\centering
\tiny
\resizebox{\columnwidth}{!}{%
\begin{tabular}{ccccc}
\hline
& \textbf{Direct Test} 
& \textbf{ActMAD} 
& \textbf{WHW} 
& \textbf{Ours} \\ \hline
\textbf{200mm} 
& 15.30 & 24.81 & 24.90 & \textbf{28.05} \\
\textbf{100mm} 
& 27.37 & 35.55 & 32.65 & \textbf{36.93} \\
\textbf{75mm}  
& 31.88 & 38.61 & 38.71 & \textbf{40.06} \\ \hline
\end{tabular}%
}
\caption{Transformer-based detector (Swin-T) on KITTI rain.}
\label{tab:table_re_4}
\end{table}
\begin{table}[hbt!]
\centering
\resizebox{\columnwidth}{!}{%
\begin{tabular}{ccccccc}
\hline
\textbf{} & \textbf{Direct Test} & \textbf{Buffer TTA} & \textbf{Pruning TTA} & \textbf{ActMAD} & \textbf{WHW} & \textbf{Ours} \\ \hline
\textbf{Gaussian} & 8.87  & 11.69 & 31.60  & 40.14 & 23.15          & \textbf{43.43} \\
\textbf{Jpeg}     & 65.66 & 70.65 & 66.02 & 62.65 & \textbf{72.45} & 70.59          \\
\textbf{Motion}   & 47.81 & 51.55 & 63.49 & 63.74 & 64.49          & \textbf{72.37} \\
\textbf{Defocus}  & 52.6  & 57.14 & 63.27 & 61.81 & 66.77          & \textbf{70.7}  \\ \hline
\end{tabular}%
}
\caption{Evaluation on non-weather corruptions(KITTI).}
\label{tab:table_re_5}
\end{table}

\noindent{\textbf{Scope and Generalization.}}

We evaluate the proposed method on a Swin-T Transformer-based detector, where BN is replaced with LN and MLP-based adapters are used (Tab.~\ref{tab:table_re_4}). The results show consistent improvements over baseline methods across all shift severities. The ACDC benchmark already includes compound domain shifts, such as sensor differences between datasets and diverse appearance variations (e.g., brightness, blur, and contrast). To further assess generalization, we additionally evaluate on non-weather corruptions, including Gaussian noise, JPEG compression, motion blur, and defocus blur (Tab.~\ref{tab:table_re_5}), where our method achieves the best or comparable performance.


\begin{table}[hbt!]
\centering
\resizebox{\columnwidth}{!}{%
\begin{tabular}{ccccccc}
\hline
& \textbf{Direct Test} 
& \textbf{Buffer TTA} 
& \textbf{Pruning TTA} 
& \textbf{ActMAD} 
& \textbf{WHW} 
& \textbf{Ours} \\ \hline
\textbf{30m}  
& 12.37 & 14.54 & 23.89 & 28.31 & 14.79 & \textbf{33.04} \\
\textbf{375m} 
& 69.57 & 73.70 & 70.60 & 65.55 & 72.73 & \textbf{74.97} \\
\textbf{750m} 
& 74.67 & 78.09 & 74.89 & 68.07 & 77.65 & \textbf{78.58} \\ \hline
\end{tabular}%
}
\caption{Extended severity evaluation on KITTI fog.}
\label{tab:table_re_ex_capa}
\end{table}
\noindent{\textbf{Performance Across Extended Visibility Ranges.}}

We extend the evaluation to a broader range of visibility conditions beyond the primary range (Tab.~\ref{tab:table_re_ex_capa}). Under milder conditions ($\geq$375m), the direct test performance is already close to the clean setting, leaving limited room for improvement. The results show that subtractive approaches contribute more under severe shifts (30m), while additive approaches are more effective under milder shifts (375m, 750m), consistent with the complementary pattern observed in Fig. 1 (main paper). Our method achieves the best performance across all visibility levels.

\begin{figure*}[hbt!]
    \centering
    \includegraphics[width=\textwidth]{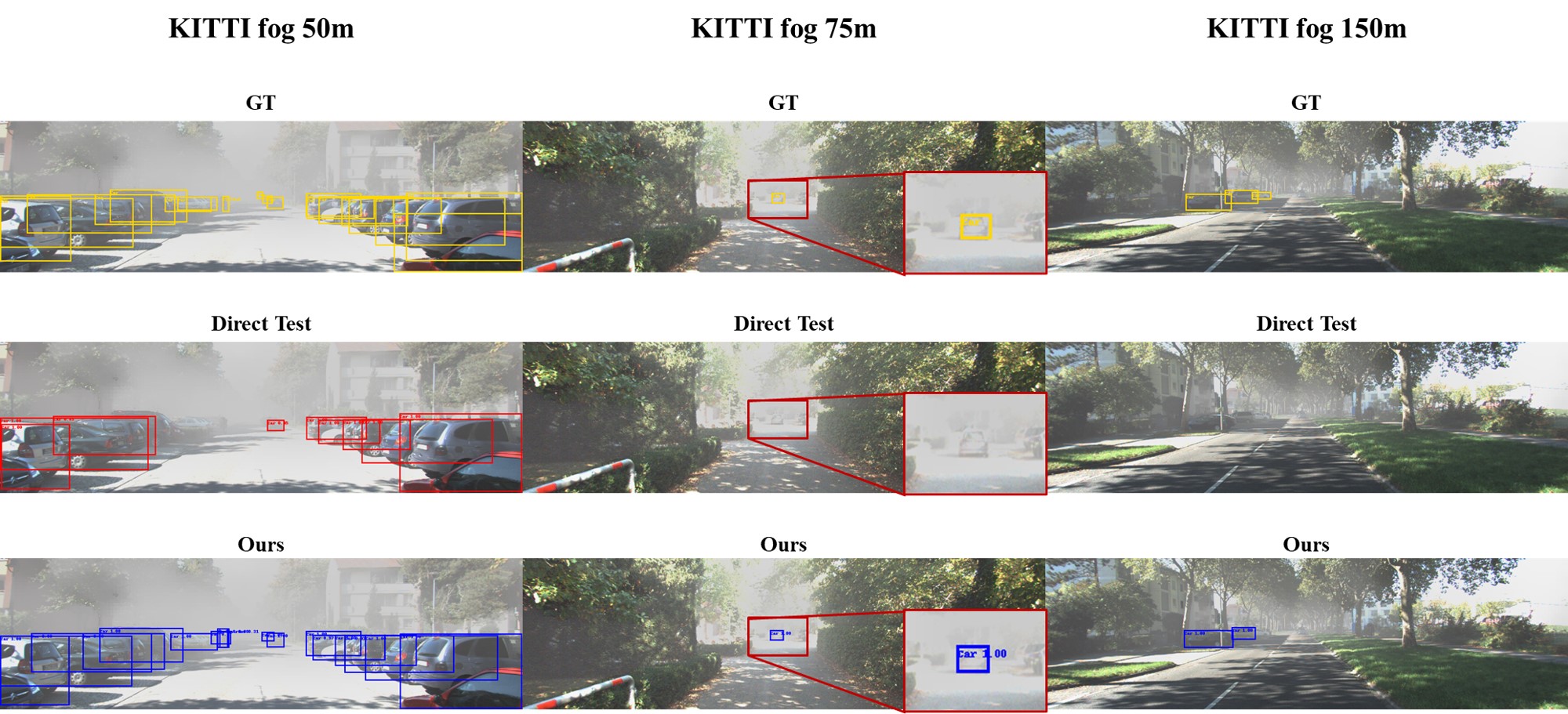}
    \caption{\textbf{Qualitative results on KITTI fog.} Object detection results visualized with bounding boxes across different fog severities (50m, 75m, 150m). We compare ground truth (GT), Direct Test (source model without adaptation), and our CD-Buffer results, demonstrating effective adaptation across all severity levels.}
    \label{fig6.kitti_fog_qual}
\end{figure*}

\begin{figure*}[hbt!]
    \centering
    \includegraphics[width=\textwidth]{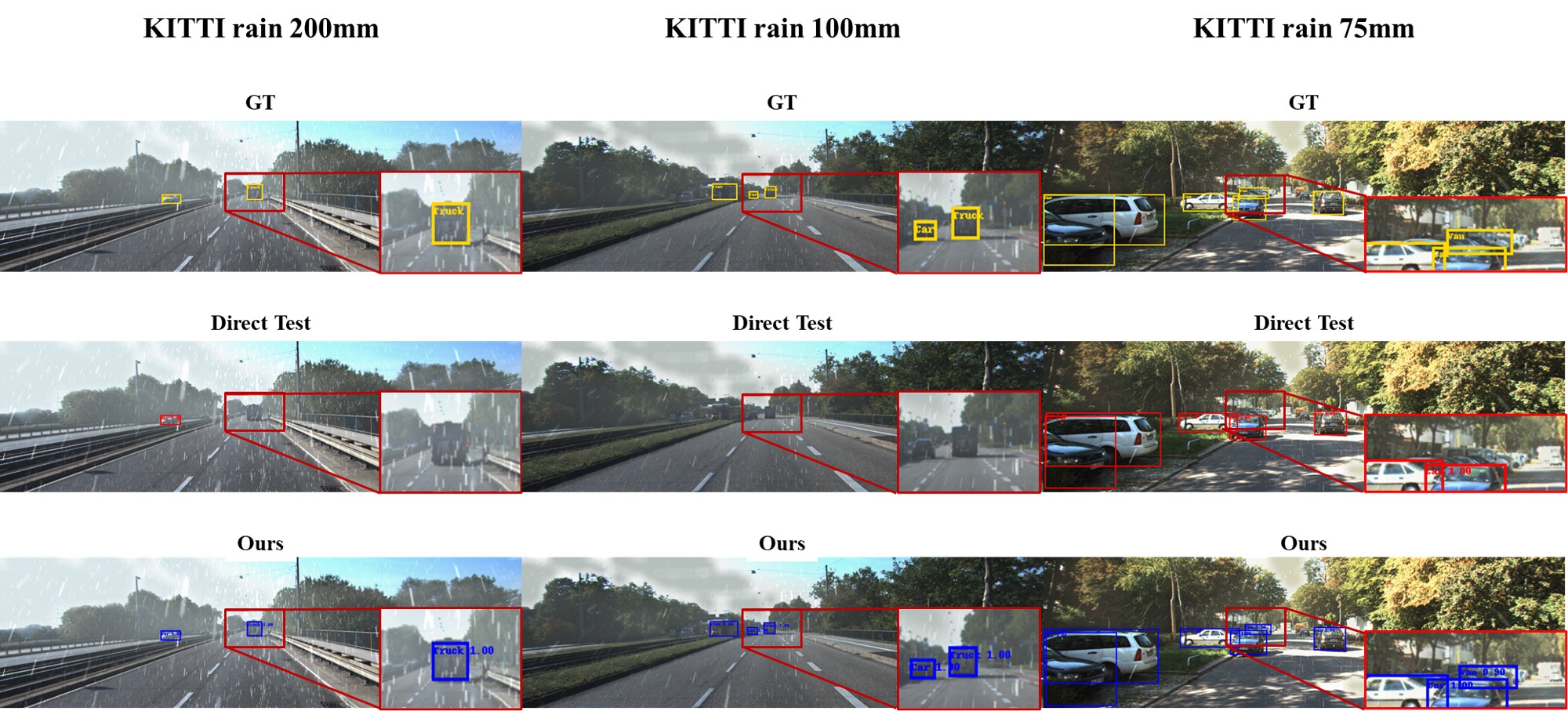}
    \caption{\textbf{Qualitative results on KITTI rain.} Object detection results visualized with bounding boxes across different rain severities (200mm, 100mm, 75mm). We compare ground truth (GT), Direct Test (source model without adaptation), and our CD-Buffer results, demonstrating effective adaptation across all severity levels.}
    \label{fig7}
\end{figure*}

\begin{figure*}[hbt!]
    \centering
    \includegraphics[width=\textwidth]{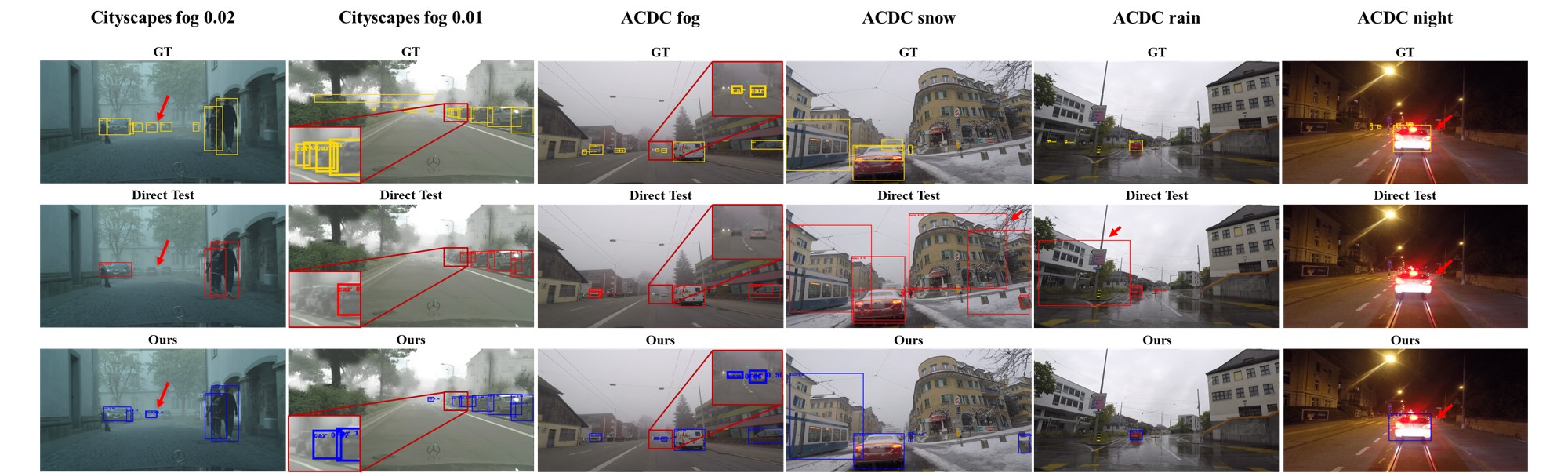}
    \caption{\textbf{Qualitative results on Cityscapes fog and ACDC.} Object detection results visualized with bounding boxes. For Cityscapes, we show fog severities (0.02, 0.01), and for ACDC, we present diverse conditions including fog, snow, rain, and night. We compare ground truth (GT), Direct Test (source model without adaptation), and our CD-Buffer results, demonstrating effective adaptation across all conditions.}
    \label{fig8}
\end{figure*}


\end{document}